\documentclass{article}

\usepackage{microtype}
\usepackage{graphicx}
\usepackage{subcaption}
\usepackage{booktabs} % for professional tables

\usepackage{hyperref}

\usepackage[preprint]{icml2026}

\usepackage{amsmath}
\usepackage{amssymb}
\usepackage{mathtools}
\usepackage{amsthm}

\definecolor{green}{HTML}{006400}
\definecolor{red}{HTML}{ff0000}

\usepackage{hyperref}
\usepackage{url}
\usepackage{graphicx}
\usepackage{amsmath}
\usepackage{CJKutf8}
\usepackage{enumitem}
\usepackage{subcaption}
\usepackage{xspace}
\usepackage{ifthen}
\usepackage{xcolor}
\usepackage{amsmath}
\usepackage{mathtools}
\usepackage{tabularx}
\usepackage{stackengine}
\usepackage{bbm}
\setstackgap{S}{1pt}
\usepackage{tikz}
\usetikzlibrary{tikzmark,arrows.meta}

\tikzset{
  relarrowA/.style={->, >=Stealth, line width=0.2pt, bend left=55},
}

\usepackage[T1]{fontenc}
\catcode`\"=12
\usepackage{graphicx}
\usepackage{amsmath}
\usepackage{caption}
\usepackage{subcaption}
\usepackage{xspace}
\usepackage{array}
\usepackage{comment}
\usepackage{lineno}
\usepackage{wrapfig}

\usepackage{CJKutf8}

\makeatletter
\renewcommand{\sectionautorefname}{\S\@gobble}
\renewcommand{\subsectionautorefname}{\S\@gobble}
\renewcommand{\subsubsectionautorefname}{\S\@gobble}
\renewcommand{\appendixautorefname}{Appendix \@gobble}
\makeatother

\newcommand{\lm}[1]{\textsc{#1}}

\definecolor{purp}{HTML}{791f87}
\definecolor{green}{HTML}{006400}

\newcommand{\whitespace}{\,\textvisiblespace\,}

\DeclareRobustCommand{\newline}{%
  \begingroup\normalfont
  \includegraphics[height=9pt]{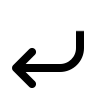}%
  \endgroup
}

\definecolor{tokencolor}{HTML}{AA3377}
\definecolor{charcolor}{HTML}{31688E}
\newcommand{\textcharcolor}[1]{\textcolor{charcolor}{#1}}
\newcommand{\texttokencolor}[1]{\textcolor{tokencolor}{#1}}

\newcommand{\chars}[1]{%
  {\color{charcolor}{%
    \text{``}#1\text{''}}}%
}

\newcommand{\token}[1]{%  
  {\color{tokencolor}{%
    \texttt{#1}}}%
}

\newcommand{\annot}[2]{
    \stackon{\text{#2\strut}}{\scriptsize\text{#1\strut}}%
}

\DeclareMathOperator{\encode}{encode}

\usepackage{titlesec}

\usepackage[table]{xcolor}
\usepackage{pgf}
\newcommand{\heat}[1]{%
  \pgfmathsetmacro{\percent}{round(min(#1,100))}% round and cap at 100
  \edef\temp{\noexpand\cellcolor{red!\percent!white}{#1}}%
  \temp
}

\usepackage[capitalize,noabbrev]{cleveref}

\theoremstyle{plain}

\theoremstyle{definition}

\theoremstyle{remark}

\usepackage[textsize=tiny]{todonotes}

\icmltitlerunning{\textit{Are you going to finish that?} A Practical Study of the Partial Token Problem}

\begin{document}
\begin{CJK*}{UTF8}{gbsn}

\twocolumn[
  \icmltitle{\textit{Are you going to finish that?}\\A Practical Study of the Partial Token Problem}



  \icmlsetsymbol{equal}{*}

  \begin{icmlauthorlist}
    \icmlauthor{Hao Xu}{uw}
    \icmlauthor{Alisa Liu}{uw}
    \icmlauthor{Jonathan Hayase}{uw}
    \icmlauthor{Yejin Choi}{nvidia}
    \icmlauthor{Noah A. Smith}{uw,aiTwo}
  \end{icmlauthorlist}

  \icmlaffiliation{uw}{Paul G. Allen School of Computer Science and Engineering, University of Washington}
  \icmlaffiliation{nvidia}{Nvidia}
  \icmlaffiliation{aiTwo}{Allen Institute for AI}

  \icmlcorrespondingauthor{Hao Xu}{xuhao510@cs.washington.edu}

  \icmlkeywords{Machine Learning, ICML}

  \vskip 0.3in
]



\printAffiliationsAndNotice{}  % no special notice (required even if empty)

\begin{abstract}
Language models (LMs) are trained over sequences of tokens, whereas users interact with LMs via text.
This mismatch gives rise to the \textit{partial token problem}, which occurs when a user ends their prompt in the middle of the expected next-token, leading to distorted next-token predictions.
Although this issue has been studied using arbitrary character prefixes, its prevalence and severity in \textit{realistic} prompts respecting word boundaries remains underexplored.
In this work, we identify three domains where token and ``word'' boundaries often do not line up: languages that do not use whitespace, highly compounding languages, and code.
In Chinese, for example, up to 25\% of word boundaries do not line up with token boundaries, making even natural, word-complete prompts susceptible to this problem.
We systematically construct \textit{semantically natural} prompts ending with a partial tokens; in experiments, we find that they comprise a serious failure mode: frontier LMs consistently place three \textit{orders of magnitude} less probability on the correct continuation compared to when the prompt is ``backed-off'' to be token-aligned.
This degradation does not diminish with scale and often worsens for larger models.
Finally, we evaluate inference-time mitigations to the partial token problem and validate the effectiveness of recent exact solutions.
Overall, we demonstrate the scale and severity of probability distortion caused by tokenization in realistic use cases, and provide practical recommentions for model inference providers.
\end{abstract}

\section{Introduction}\label{sec:introduction}
Users of language models (LMs) generally interface with LMs via text, providing a \textcharcolor{\textbf{string of characters}} as prompt and expecting a string continuation as output.
Under the hood, this is accomplished by using a tokenizer to segment text into a \texttokencolor{\textbf{sequence of tokens}}, and prompting LMs trained over tokens to predict the next-token, which is then decoded back into text.
While this generally occurs seamlessly and unbeknownst to the user, undesirable consequences of the mismatch appear when users (potentially unknowingly) provide a prompt that ends with the \textit{prefix} of a valid token, which then causes the model to assign unexpectedly \textit{low} probability to the completion of that token.

This fundamental issue is known as the \textit{partial token problem} (PTP; \autoref{sec:background}).\footnote{In other works, this issue has been called the ``prompt boundary problem'' or ``tokenization bias,'' but we choose to describe it as the \textit{partial token problem} both to highlight the role of tokenization and to accurately reflect that the problem does not \textit{only} occur at the prompt boundary. (See next footnote)}
It underlies (for example) the standard advice to users to not end prompts with a trailing whitespace, as the whitespace would be a partial token relative to the many tokens in the vocabulary that start with a whitespace.
While the PTP has been extensively documented in prior work \citep{lundberg-2023-art, ribeiro2023guidance, phan-etal-2025-exact} and many methods have been proposed to address it \citep{athiwaratkun-etal-2024-token, vieira-etal-2025-from, phan-etal-2025-exact, turaga-2025-character, hayase-etal-2025-sampling}, these works generally illustrate the PTP using arbitrary character prefixes of the text; as a result, the extent to which the PTP affects natural use cases remains unknown.

\begin{figure*}[t]
    \centering
    \includegraphics[width=\linewidth]{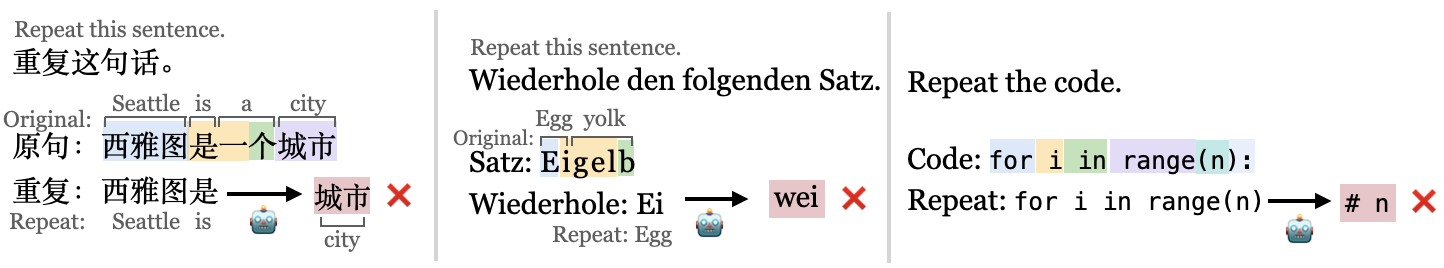}
    \caption{Example model failures at performing a simple repeat-after-me task when given prompts that align with natural word boundaries but end with a partial token.
    \textbf{Left:} In Chinese, because \token{是一个}\,(\textit{is a}) is a single token, the model does not generate the correct continuation \chars{一个}\,(\textit{a}) given \chars{是} (\textit{is}). 
    \textbf{Middle:} In German, the model does not generate \chars{gelb} (\textit{yolk}) given \chars{\whitespace Ei} (\textit{egg}). \textbf{Right:} In code, the model does not generate the colon \chars{:} given \chars{)} because \token{):} is a single token. Model continuations are shown for \lm{Qwen3-4B}.}
    \label{fig:boundary-example}
\end{figure*}

In this work, we make the first attempt to demonstrate the severity of the PTP in natural and realistic prompts, including both its prevalence and the level of distortion it wreaks on next-token probability distributions.
We identify three domains where whitespace does not reliably separate semantic or syntactic units (see illustrations in Fig.~\ref{fig:boundary-example}), where even prompts with complete word endings are susceptible (\autoref{sec:misalignment}).

\begin{itemize}[leftmargin=12pt]
    \item \textbf{Logographic writing systems.} Unlike English, where words are typically separated by whitespace, words in languages such as Chinese and Japanese are not delimited.
    Tokenizers can therefore merge characters across word boundaries, and as a result, token and word boundaries often do not align. 
    For instance, \emph{是/一个} (\emph{is} / \emph{a}) is tokenized as the single token $\langle\token{是一个}\rangle$ in the \lm{Qwen3} tokenizer, meaning that a prompt that ends with the complete word \chars{是} (\textit{is}) would be affected by the PTP.
    \item \textbf{Languages with highly productive compounding}.
    In German, words are often built by combining two or more complete words.
    Thus, a tokenizer can segment compound words in ways that do not respect morphological boundaries.
    For example, \textit{Eigelb} (\emph{egg yolk}), which comes from \emph{Ei} (\emph{egg}) and \emph{gelb} (\textit{yellow}), is tokenized
    as $\langle\token{\whitespace E}, \token{igel}, \token{b}\rangle$, so a prompt ending with \chars{\whitespace Ei} would similarly be affected by the PTP.
    \item \textbf{Code completion.} Code often contains long identifiers or consecutive punctuation that are merged into single tokens. For example, in the Python main function definition \chars{def main():}, the final sequence of punctuation $\langle\token{():}\rangle$ is a single token. As a result, the PTP affects prompts that ends with the punctuation \chars{()}.
\end{itemize}

Then, we systematically construct prompts that align with natural word boundaries but end with a partial token relative to the expected continuation, in order to create controlled experimental conditions that expose the PTP (\autoref{sec:experiments}).
We achieve this by constructing a ``repeat-after-me'' task in which the model is asked to repeat a given sentence with the target text partially provided.
We find across a wide range of models that the PTP degrades model accuracy dramatically, by 60\% -- 95\%.
In Chinese, the probability assigned to the correct next-token drops by four orders of magnitude!
Even more, this behavior does not improve with scaling.

Finally, we investigate the effectiveness of proposed solutions to the PTP.
Token healing \citep{ribeiro2023guidance} is a simple heuristic approach that ``backs off'' the prompt by removing a token from the prompt ending.
More recently, some works have proposed \textit{exact} solutions to the PTP \citep{vieira-etal-2025-from, phan-etal-2025-exact, turaga-2025-character, hayase-etal-2025-sampling}, which claim to preserve the probability distribution of the LM at the level of text.
For instance, \citet{hayase-etal-2025-sampling} constructs a tree of all possible token sequences that ``cover'' the provided prompt, then samples a path from root to leaf.
We find that this approach is effective, improving performance on the trivial repeat-after-me task to 100\%, while heuristic approaches like token healing have more mixed performance.
% Because ByteSampler is also efficient (using $\sim$1 additional forward pass on average to move past the prompt boundary, after which sampling can resume in the normal way), we recommend that all deployments of tokenizer-based LMs use an inference-time approach to correct for PTP distortion at the end of the user prompt.

Overall, our work highlights how a fundamental problem caused by tokenization surfaces in realistic use cases, and provides new evaluations of a recent solution.

\section{Background on partial token problem}\label{sec:background}

% \alisa{possible approach to formalizing the PTP below. not sure if this makes sense because right now I'm separating when PTP happens vs. when it doesn't, but I think PTP is technically always present because every token excludes the possibility of some next-tokens under canonical tokenization. e.g., \texttt{\whitespace the} excludes the next-token \texttt{re} (because \texttt{\whitespace there} is a token), and that technically distorts the probability distribution for all the other possible next-tokens even when \texttt{re} isn't what you care about.}

% Language models are internally models over tokens, whereas users interact with LMs via string prompts and continuations.

The task of language modeling is to define a probability
\[P(\operatorname{\textcharcolor{continuation}} \mid \operatorname{\textcharcolor{prompt}})\]
for given \(\operatorname{\textcharcolor{prompt}}\) over all possible \(\operatorname{\textcharcolor{continuation}}\) strings. 
In practice, tokenizer-based LMs model this by encoding the text as a sequence of tokens and autoregressively calculating the probability of each subsequent token,
\begin{multline}\label{eq:1}
    P(\texttokencolor{c_1},\texttokencolor{...}, \texttokencolor{c_n}\mid \texttokencolor{p_1},\texttokencolor{...}, \texttokencolor{p_m}) \\
    = \prod_{i=1}^n P(\texttokencolor{c_i}\mid \texttokencolor{p_1},...,\texttokencolor{p_m},\texttokencolor{c_1},...,\texttokencolor{c_{i-1}})
\end{multline}
% \alisa{indexing in RHS is a little cursed :/}
where \(\langle\texttokencolor{p_1},\texttokencolor{...}, \texttokencolor{p_m}\rangle\leftarrow \encode(\operatorname{\textcharcolor{prompt}})\) and \(\langle\texttokencolor{c_1},\texttokencolor{...}, \texttokencolor{c_n}\rangle\leftarrow\encode(\operatorname{\textcharcolor{continuation}})\).
Because LMs are trained only on valid token sequences, namely, token sequences in the output space of \(\encode\), the prediction of \autoref{eq:1} is only in-distribution with training when the concatenation of the prompt and continuation, tokenized individually, \(\langle\texttokencolor{p_1},\texttokencolor{...}, \texttokencolor{p_m}, \texttokencolor{c_1},\texttokencolor{...}, \texttokencolor{c_n}\rangle\), is also a valid token sequence, or equivalently,
\begin{multline}\label{eq:condition}
    \encode(\operatorname{\textcharcolor{prompt}}) + \encode(\operatorname{\textcharcolor{continuation}})= \\
    \encode(\operatorname{\textcharcolor{prompt}} + \operatorname{\textcharcolor{continuation}})
\end{multline}
where we use $+$ to denote both string and list concatenation.
Note that \autoref{eq:condition} is not true, in general, because the ending of $\operatorname{\textcharcolor{prompt}}$ (on the left side) forces a token boundary that may or may not be present when $\operatorname{\textcharcolor{prompt}}$ and $\operatorname{\textcharcolor{continuation}}$ are tokenized together (right side).
When \autoref{eq:condition} is not satisfied,  the \textit{partial token problem} (PTP) may arise. 
In this case, there is a token $\texttokencolor{T}$ from \(\encode(\operatorname{\textcharcolor{prompt}} + \operatorname{\textcharcolor{continuation}})\) that bridges the prompt-continuation boundary, but is not present (in that position) in \(\encode(\operatorname{\textcharcolor{prompt}}) + \encode(\operatorname{\textcharcolor{continuation}})\).
% we denote $\textcharcolor{s_1}$ as the string portion in the prompt and $\textcharcolor{s_2}$ the portion in the continuation. 
% Further, let $\texttokencolor{t_1}=\encode(\textcharcolor{s_1})$ (tokens from the ending of $\langle\texttokencolor{p_1},\texttokencolor{...}, \texttokencolor{p_m}\rangle$) and $\texttokencolor{t_2}=\encode(\textcharcolor{s_2})$ (tokens from the beginning of $\langle\texttokencolor{c_1},\texttokencolor{...}, \texttokencolor{c_n}\rangle$) be the corresponding token sequences.
% \alisa{ok, there's a really subtle thing that's wrong here that I don't know how to deal with, which is that $\texttokencolor{t_2}$ might not actually be a prefix of the tokenized continuation $\langle\texttokencolor{c_1},\texttokencolor{...}, \texttokencolor{c_n}\rangle$. this is because the new token boundary between $s_1$ and $s_2$ might change whether there is still a token boundary after $s_2$. but the problem is that our evals may have this assumption baked in, so I need to check with Hao.}
% Note that $\textcharcolor{s_1}+\textcharcolor{s_2}=\operatorname{decode}(\texttokencolor{T})$ but $\texttokencolor{t_1} + \texttokencolor{t_2} \neq \texttokencolor{T}$.
% Note that $\texttokencolor{T}$ and $\texttokencolor{t_1} + \texttokencolor{t_2}$ decode to the same string $\textcharcolor{s_1}+\textcharcolor{s_2}$, but $\texttokencolor{T}$ is the canonical tokenization whereas $\texttokencolor{t_1} + \texttokencolor{t_2}$ is the tokenization forced by the prompt boundary.
Formally, a prompt suffers from the PTP when the \textit{probability} of the sequence $\encode(\operatorname{\textcharcolor{prompt}}) + \encode(\operatorname{\textcharcolor{continuation}})$ (which contains a token boundary forced by the prompt ending) differs from that of \(\encode(\operatorname{\textcharcolor{prompt}} + \operatorname{\textcharcolor{continuation}})\) (the canonical tokenization of the entire string).
% \alisa{does this seem reasonable...} \nascomment{yes but I don't love + and -- as operators on strings.  if you use them, I think you should explain up front that you are doing so.}

% Let us now use a practical example to illustrate the PTP with \lm{Llama3-3.2-1B}.
% Suppose we have \(\operatorname{\textcharcolor{prompt}}=\chars{  University of Washingto}\), and \(\operatorname{\textcharcolor{continuation}}=\chars{n}\). \alisa{I'll change the example later since this probably breaks anonymity}
% In this case, \autoref{eq:condition} is not satisfied:

% \[\encode(\operatorname{\textcharcolor{prompt}}) + \encode(\operatorname{\textcharcolor{continuation}}) = \langle \token{\whitespace University}, \token{\whitespace of}, \token{\whitespace Washing}, \token{to}, \token{n}\rangle\]

% \[\encode(\operatorname{\textcharcolor{prompt}} + \operatorname{\textcharcolor{continuation}}) = \langle\token{\whitespace University}, \token{\whitespace of}, \token{\whitespace Washington}\rangle\]

A practical (if contrived) English example illustrates the PTP with \lm{Llama3-3.2-1B}.
Suppose we have \(\operatorname{\textcharcolor{prompt}}=\) \chars{\whitespace natural\whitespace language\whitespace processin}, and \(\operatorname{\textcharcolor{continuation}}=\chars{g}\).
In this case, \autoref{eq:condition} is not satisfied, as
\begin{multline*}
\encode(\operatorname{\textcharcolor{prompt}} + \operatorname{\textcharcolor{continuation}}) =\\
\langle\token{\whitespace natural}, \token{\whitespace language}, \token{\whitespace processing}\rangle
\end{multline*}
\begin{multline*}
\encode(\operatorname{\textcharcolor{prompt}}) + \encode(\operatorname{\textcharcolor{continuation}}) =\\
\langle \token{\whitespace natural}, \token{\whitespace language}, \token{\whitespace process}, \token{in}, \token{g} \rangle
\end{multline*}

% In this case, we see that the token \(T=\texttt{\whitespace Washington}\) bridges the prompt-continuation boundary, with \(s_1=\texttt{"Washingto"}\) and \(s_2=\texttt{"n"}\).

Here, the token \(\texttokencolor{T}=\token{\whitespace processing}\) bridges the prompt-continuation boundary.
To verify the effect of the PTP, we observe that, under the model' s distribution:\[P(\encode(\operatorname{\textcharcolor{prompt}} + \operatorname{\textcharcolor{continuation}})) = 4.77 \times 10^{-8}\]\[P(\encode(\operatorname{\textcharcolor{prompt}}) + \encode(\operatorname{\textcharcolor{continuation}})) = 1.58 \times 10^{-15}\]
which differ substantially despite corresponding to the same string.\footnote{In order to set up notation for the experiments in our paper, we have focused on the scenario where the PTP occurs at the end of the prompt. 
However, the PTP can also occur at the end of the continuation. 
For instance, the string \chars{\whitespace processin} given \chars{\whitespace natural\whitespace language} has lower likelihood ($1.35 \times 10^{-5}$) under \lm{Llama3.2} than \chars{\whitespace processing} given the same context ($0.788$), despite \chars{\whitespace processin} being a substring of \chars{\whitespace processing}. This is because the likelihood of the token sequence $\encode(\chars{\whitespace processin})$ actually \textit{excludes} the probability of the string $\chars{\whitespace processing}$, for which the single token $\token{\whitespace processing}$ would be used.}
In particular, the probability of token \token{g} when conditioned on \chars{\whitespace natural\whitespace language\whitespace processin} is only 0.002, even though it is overwhelmingly the most reasonable continuation.
Intuitively, this is because the LM has learned that the string \chars{\whitespace processing} is always encoded with the single token $\langle\token{\whitespace processing}\rangle$. Thus the context $\langle\token{\whitespace process}, \token{in}\rangle$ directly indicates that \chars{g} is not the correct continuation (otherwise \token{\whitespace processing} would have been used!).

Fortunately, the PTP is generally avoidable in English and other whitespace-delimited languages.
% languages that use whitespace-delimited words.
This is because the \textit{pretokenization step} in tokenizer training splits the training text on whitespace, preventing tokens bridging whitespace from being learned; in particular, since splits conventionally occur to the \textit{left} of whitespace, learned tokens can only have \textit{leading} (not trailing) whitespace.
By ending prompts with complete words and no trailing whitespace, user prompts are thus guaranteed to line up with a token boundary.
% The PTP is the reason why users are universally advised to not end their prompt with a trailing whitespace, as this would be a ``partial token'' with respect to many tokens and empirically leads to much degraded results.

% Tokenization segments text, as a stream of bytes, into tokens in the LM vocabulary. 
% Traditionally, tokenization is deterministic.

% This is because the end of the prompt forces a token boundary.
% \alisa{talk about whitespace pretokenization, making PTP not a thing for English tokenization}

% The partial token problem (PTP) occurs whenever the prompt ends on a prefix of what could otherwise be a single token.

% While some works have introduced test-time approaches for overcoming the PTP, our work focuses on rigorously quantifying the impacts of this problem in realistic use cases.

Nonetheless, many methods have been proposed to address the PTP.
For instance, the heuristic technique of token healing \citep{ribeiro2023guidance, dagan-etal-2024-getting, athiwaratkun-etal-2024-token} ``backs off'' the prompt by removing one or more tokens from the prompt ending, then sampling a continuation that is constrained to match the removed text.
In contrast, exact methods preserve the probability distribution over text of the original LM \citep{vieira-etal-2025-from, phan-etal-2025-exact, turaga-2025-character, hayase-etal-2025-sampling}.
In both cases, these works generally evaluate the PTP using arbitrary character prefixes of the text, which may not reflect realistic user prompts.
In this work, we instead study how commonly the PTP arises in practice and whether the probability distortion persists when prompt endings align with natural word or syntactic boundaries.

\begin{table*}[h]
    \centering
    \caption{\textbf{There is a high rate of misalignment between token units and semantic/syntactic units} when using off-the-shelf tokenizers to encode Chinese, German, and code data, where zero misalignment means every semantic/syntactic boundary lines up with a token boundary.
    Misalignment creates the potential for the partial token problem to affect even natural prompts that respect word boundaries.
    % \alisa{TODO: remove vocab column under Chinese}
    }
    \begin{tabular}{l |r|r|rrrrrr}
        \toprule
        & \textbf{Chinese} & \textbf{German} & \multicolumn{6}{c}{\textbf{Code}} \\
        & \textbf{Wiki} & \textbf{Wiki} 
        & \textbf{Python} & \textbf{Java} & \textbf{JS} 
        & \textbf{Go} & \textbf{PHP} & \textbf{Ruby} \\
        \midrule
        \lm{Qwen 3}
        & \heat{19.96} & \heat{6.51}
        & \heat{68.08} & \heat{65.31} & \heat{61.35}
        & \heat{55.75} & \heat{52.52} & \heat{62.41} \\
        
        \lm{Deepseek-V3}
        & \heat{23.43} & \heat{4.38}
        & \heat{63.55} & \heat{60.66} & \heat{57.90}
        & \heat{52.07} & \heat{50.09} & \heat{55.61} \\
        
        \lm{Hunyuan}
        & \heat{24.92} & \heat{37.20}
        & \heat{66.28} & \heat{63.53} & \heat{59.78}
        & \heat{53.23} & \heat{51.32} & \heat{59.07} \\
        
        \lm{Llama 3}
        & \heat{24.50} & \heat{37.84}
        & \heat{68.08} & \heat{65.31} & \heat{61.35}
        & \heat{55.75} & \heat{52.52} & \heat{62.41} \\
        
        \lm{OLMo 2}
        & \heat{14.88} & \heat{8.09}
        & \heat{68.08} & \heat{65.31} & \heat{61.35}
        & \heat{55.75} & \heat{52.52} & \heat{62.41} \\
        
        \lm{Mistral NeMo}
        & \heat{18.10} & \heat{5.46}
        & \heat{65.67} & \heat{62.05} & \heat{59.33}
        & \heat{52.86} & \heat{50.99} & \heat{57.87} \\
        
        \lm{Gemma}
        & \heat{14.12} & \heat{3.40}
        & \heat{25.11} & \heat{24.90} & \heat{23.81}
        & \heat{16.21} & \heat{32.77} & \heat{16.98} \\
        \bottomrule
    \end{tabular}

    \label{tab:misalignment}
\end{table*}

\section{Misalignment between tokens and ``words''}\label{sec:misalignment}

In this section, we motivate our study of the partial token problem by quantifying how often word and token boundaries do not align in natural text.
We define the \textbf{misalignment rate} as the proportion of word boundaries (or syntactic boundaries, in the case of code) that do not line up with a token boundary.
When misalignment is high, even prompts with complete word endings are susceptible to the PTP.
We consider three domains where whitespace does not reliably separate semantic or syntactic units.

\paragraph{Chinese} We sample 1,000 entries from \href{https://huggingface.co/datasets/fjcanyue/wikipedia-zh-cn}{Chinese Wikipedia} and obtain word boundaries with the off-the-shelf segmentor \href{https://github.com/fxsjy/jieba}{\texttt{Jieba}}, a widely-used Chinese word segmentor based on a prefix dictionary. %\alisa{how did you download Chinese Wikipedia? we probably still need to cite something here, e.g., the website link or dataset that you used}

\paragraph{German} We sample 1,000 entries from \href{https://huggingface.co/datasets/jonas-is-coding/german-wikipedia-articles}{German Wikipedia} and obtain word boundaries with \href{https://github.com/dtuggener/CharSplit}{\texttt{CharSplit}}, an $n$-gram-based compound splitter built specifically for German.

\paragraph{Code} 
We sample 200 snippets for each of six programming languages from the CodeXGLUE dataset \citep{husain2019codesearchnet}.
Since each punctuation character in code usually delimits some syntactic boundary (e.g., the closing a parenthesis), here we consider the misalignment rate to be the proportion of right-boundaries of punctuation characters that lie in the middle of a token.

Shown in \autoref{tab:misalignment}, off-the-shelf tokenizers produce a high misalignment rate across the board. 
In Chinese, 14\% -- 25\% of word boundaries do not lie on a token boundary.
% Prompts ending with these words would receive a distorted next-token prediction, due to forcing a new token boundary not originally present in that position.
We observe that this misalignment rate is not necessarily tied to the number of Chinese tokens in the vocabulary: \lm{Llama 3} and \lm{Hunyuan}, with 4,301 and 45,131 Chinese tokens, respectively, exhibit similar misalignment rates (24.5\% and 24.9\%). 
% The difference may come from vocabulary composition: \lm{Llama 3} includes many tokens beginning with punctuation (e.g., ``，'' or ``。''), which increases the likelihood of word boundaries falling inside tokens. 
% \alisa{Hao elaborate here}.
% \alisa{@Hao how quickly can you count the number of Chinese tokens in the vocabularies of each of these tokenizers?}
% \hao{I removed Chinese vocab size because it is actually not related.. like Llama only has 4301 Chinese tokens but it has the second highest misalignment rate. I also checked the average token length but it seems not quite related too. I guess it can depend on which characters get merged, e.g. Llama has a lot of tokens starting with punctuations.}\alisa{thanks! IMO that's really interesting. can we make a comment then that it's NOT related to the proportion of Chinese tokens?}
% \hao{and i think the misalignment rate can varies largely depending on the text being tested. many long tokens in Hunyuan, for example, comes from Chinese news, and if we are testing on those corpora then its misalignment would be really high.}\alisa{can you summarize this analysis a little bit? it's okay if it's qualitative and hand-wavy, but otherwise someone reading this will naturally wonder if it's all about the number of Chinese tokens in the vocabulary}
The misalignment in German is lower, as whitespace is used in conjunction with word compounding.

For code, punctuation characters commonly lie in the middle of longer tokens, at a rate of $\geq 50\%$ across all tokenizers and programming languages, except for \lm{Gemma}'s tokenizer at 16.98\%.
The high misalignment rate is unsurprising as punctuation characters commonly occur contiguously in code, and most commonly used pretokenizers allow punctuation to be merged together freely.
% The exception is \lm{Gemma}, whose more restrictive pretokenization results in a lower misalignment rate.
Note that we focus on syntactic boundaries in code for consistency with the natural language setting, but in the popular use case of code autocompletion, users may pause their typing anywhere, including in the middle of function names or identifiers that do not represent a syntactic boundary.\footnote{In fact, Cursor describes this exact problem and calls for solutions in a \href{https://www.cursor.com/en/blog/cpc}{blog post} from January 2025.}
Thus our definition of misalignment focuses on a \textit{subset} of cases where the PTP would matter in practice.
% \autoref{tab:misalignment} presents the misalignment rate of seven frontier models on Chinese and German Wikipedia, as well as code snippets from six programming languages.
% Shown in \autoref{tab:misalignment}, tokenization produces a high misalignment rate. In Chinese, \lm{Hunyuan} has 24.92\% of word boundaries lying within a single token. In German, 

\begin{table*}[t]
    \centering
    \caption{\textbf{Example texts from our dataset}, which splits naturally occurring sentences into a prefix and continuation that exposes the partial token problem.
    The instruction (not pictured) asks the model to repeat the full sentence, and provides the sentence prefix as a partial completion.
    In each row, the \textit{word-aligned prompt} is shown on top, where we use \texttokencolor{red} to highlight the text corresponding to the token $\texttokencolor{T}$ that is truncated prematurely by the prompt ending.
    The \textit{token-aligned prompt} is shown on the bottom, which backs off the prompt to the token boundary preceding $\texttokencolor{T}$.
    The predicted continuation is shown for \lm{Qwen3-32B}.}
    \begin{tabular}{p{0.55\textwidth} p{0.16\textwidth} p{0.16\textwidth}}
    \toprule
        \textbf{Prefix} & \textbf{True cont.} & \textbf{Predicted cont.}  \\\midrule
        \(\annot{It}{它}\annot{is}{\texttokencolor{是}}\)& \(\annot{a}{\texttokencolor{一个}}\annot{island}{群岛}\) & \(\annot{island}{群岛}\)\\
        \(\annot{It}{它}\)& \(\annot{is a}{\texttokencolor{是一个}}\annot{island}{群岛}\) & \(\annot{is a}{是一个}\annot{island}{群岛}\) \\\midrule
        \(\annot{It}{它}\annot{commonly}{经常}\,\,\annot{appears}{\texttokencolor{出现}}\)&\(\annot{in}{\texttokencolor{在}}\)&\(\annot{and}{于}\)\\
        \(\annot{It}{它}\annot{commonly}{经常}\)&\(\annot{appears in}{\texttokencolor{出现在}}\)&\(\annot{appears in}{出现在}\)\\\midrule
        \annot{The}{Der} \annot{largest}{größte} \annot{part}{Tiel} \annot{of the}{des} \annot{route}{Wegverlaufs} \annot{is}{ist} \annot{with}{mit} \annot{stone}{Ste\texttokencolor{in}} & \annot{steps}{\texttokencolor{st}ufen} & \annot{plates}{platten}\\
        \annot{The}{Der} \annot{largest}{größte} \annot{part}{Tiel} \annot{of the}{des} \annot{route}{Wegverlaufs} \annot{is}{ist} \annot{with}{mit} \annot{sto...}{Ste} & \annot{ne steps}{\texttokencolor{inst}ufen}& \annot{ne steps}{instufen}\\\midrule
        \annot{The}{Der} \annot{easiest}{einfachste} \annot{way}{Weg} \annot{to}{um} \annot{an}{einen} \annot{introduction}{Einstieg}\annot{in}{in} \annot{the}{die} \annot{travel}{Re\texttokencolor{ise}} &\(\annot{literature}{\texttokencolor{l}iteratur}\) & \annot{reporter}{berichter}\\
        \annot{The}{Der} \annot{easiest}{einfachste} \annot{way}{Weg} \annot{to}{um} \annot{an}{einen} \annot{introduction}{Einstieg} \annot{in}{in} \annot{the}{die} \annot{trav...}{Re} & \annot{el literature}{\texttokencolor{isel}iteratur}& \annot{el report}{isebericht} \\ \midrule
        \texttt{def \texttokencolor{\_}} & \texttt{\texttokencolor{\_}init\_\_} & \texttt{\textcolor{gray}{\whitespace}\_(self)} \\
        \texttt{def} & \texttt{\texttokencolor{\_\_}init\_\_} & \texttt{\_\_init\_\_} \\ \midrule
        \texttt{def reverse\_words\texttokencolor{(}} & \texttt{\texttokencolor{s})} & \texttt{\_\_\_\_)} \\
        \texttt{def reverse\_words} & \texttt{\texttokencolor{(s})} & \texttt{(s)} \\
    \bottomrule
    \end{tabular}
    \label{tab:examples}
\end{table*}

\section{Experiments}\label{sec:experiments}

Having demonstrated the prevalence of disagreement between token and word boundaries in Chinese, German, and code, we now construct semantically (and syntactically) natural prompts that expose the PTP and measure the resulting effect on model predictions. \autoref{tab:examples} presents representative examples from our dataset.

% \begin{figure}
%     \centering
%     \includegraphics[width=1\linewidth]{img/experiment.png}
%     \caption{Illustration of experiment setup. \alisa{replace this caption with an actual summary of the experimental setup. slightly prefer word-aligned prompt above token-aligned prompt, because word-aligned is the case we actually care about and token-aligned is more like a baseline / reference point. also, some funny stuff is going on in the subscript of the log probs. notation is still in flux so we can update later.}} % \alisa{todo: update top left corner to match the updated representation from Figure 1}}
%     \label{fig:experiment}
% \end{figure}

% We design experiments to evaluate model robustness to the PTP across Chinese, German, and code. We construct realistic mid-token prompts that end semantically natural within a token while still have unambiguous continuation.

% We construct prompt-continuation pairs where the prompt ends at a semantic or syntactic boundary, yet not at a token boundary. I.e., it ends with \textit{a prefix} of the token that would be used if the prompt and continuation were being tokenized together.
% At the same time, we design the prompt to make the continuation highly likely.

\subsection{Prompt Construction}\label{subsec:prompt_construction}
Our goal is to create $(\operatorname{\textcharcolor{prompt}},\operatorname{\textcharcolor{continuation}})$ pairs that satisfy three criteria: (1) the prompt is semantically and syntactically natural, (2) the continuation is unambiguous, and (3) tokenizing the prompt + continuation together would produce a token spanning the prompt-continuation boundary (i.e., \autoref{eq:condition} is violated).

To satisfy conditions (1) and (3), we segment sentences into both word and token units, and identify word boundaries that do not align with token boundaries.
We then split the sentences at these positions, so that the prefix is a complete sequence of words that ends with a partial token relative to the complete sentence.
Then, in order to make the continuation unambiguous (condition (2)), we construct a ``repeat-after-me'' task where the prompt asks for a verbatim repetition of a given text, with the repetition partially provided (up to the split).

We consider a diverse suite of recent, off-the-shelf LMs, evaluating each model only on the domains and languages that it supports. 
% For Chinese, we include models trained on large-scale Chinese corpora, ensuring strong baseline performance and avoiding confounds from limited language coverage. For German and Code, we evaluate general-purpose models spanning different architecture, scales, and tokenization methods.
Note that we use the corresponding tokenizer for each model that we evaluate, meaning that the size of the dataset is slightly different for different models due to differing rates of word-token misalignment.
We describe each domain in more detail below.

\paragraph{Chinese}
% We frame the task as English-to-Chinese translation using parallel corpus FLORES \citep{nllbteam2022languageleftbehindscaling}. 
% \alisa{describe the prompt here}
% We use the parallel corpus FLORES \citep{nllbteam2022languageleftbehindscaling}, applying \texttt{Jieba} to the Chinese data for word segmentation and the tokenizer under evaluation for token segmentation.
% Prompts present the English source and request a Chinese translation, with the target translation partially provided.
We source Chinese sentences from the FLORES corpus \citep{nllbteam2022languageleftbehindscaling} and use \texttt{Jieba} to identify word boundaries.
From the set of 997 FLORES entries, we extract all cases where a word does not align with a token boundary, treating each mismatch as a separate test case (a single sentence may produce multiple test examples).
The prefix of the sentence up to the identified word boundary is then provided as a partial completion in the repetition task.
The instruction is written in Chinese, and translates to ``\textit{Repeat the following sentence.\newline Sentence: \{sentence\}\newline Repeat: \{prefix\}}''.
% The size of the test set varies for each tokenizer, as shown in \autoref{tab:result-repeat}.

\paragraph{German}
We follow the same approach as for Chinese, using German sentences from FLORES and the segmentor \texttt{CharSplit} for word segmentation. We use the same instruction prompt as above, translated to German.

\paragraph{Code}
We randomly sample 300 entries from the MBPP \citep{austin2021programsynthesislargelanguage} and extract all cases where a punctuation mark falls inside a single token, following the same procedure as for Chinese and German.
For example, in a Python function declaration \chars{def main():} where \token{():} is a single token, we would extract the prefixes \chars{def main(} and \chars{def main()}.
Again, the prompt provides a full code snippet and asks for a repeat (this time, the instruction is in English).

% \alisa{same question here. when multiple prompt truncations are available, which one do you take?}

We also experiment with a translation task from English to Chinese/German and a code completion task, where the prompts are more natural but can have multiple plausible continuations.
This setting produces the same conclusions as our main experiments and are presented in \autoref{app:translation}.

\begin{table*}[t]
    \centering
    \caption{\textbf{When facing the partial token problem, all evaluated models suffer a significant drop in prediction accuracy and probability on the correct continuation.} $\textbf{\(\Delta\) Logprob}$ measures the average difference in log probability assigned to the correct next-token given the prompt ending with a partial token (\textit{word-aligned}) versus the prompt when backed off to the nearest token boundary (\textit{token-aligned}). $\textbf{\(\Delta\) Acc (\%)}$ measures the average difference in accuracy predicting the true continuation given the word-aligned vs. token-aligned prompt. \textbf{$N$} is the size of the test dataset for different tokenizers. We find that when facing the PTP, models place at least two orders of magnitude less probability on the true continuation and predict it correctly at least 60\% less often (in absolute terms).}
    \begin{tabular}{lrrrrr}
        \toprule
        \textbf{Model} & $N$ & \textbf{\(\Delta\) Logprob} & \textbf{\(\Delta\) Acc (\%)} & \textbf{Acc$_{\text{word}}$ (\%)} & \textbf{Acc$_{\text{token}}$ (\%)} \\
        \midrule
        \multicolumn{6}{c}{\textbf{Chinese}} \\
        \lm{Qwen3-32B} & \(2304\) & \(-7.49\) & \(-78.56\) & \(15.36\) & \(93.92\) \\
        \lm{Hunyuan-4B-Pretrain}   & \(3551\) & \(-7.04\) & \(-77.64\) & \(16.36\) & \(94.00\) \\
        \midrule
        \multicolumn{6}{c}{\textbf{German}} \\
        \lm{Qwen3-32B}             & \(493\)   & \(-6.61\) & \(-73.02\) & \(16.43\) & \(89.45\) \\
        \lm{Llama-3.1-8B}          & \(513\)   & \(-4.18\) & \(-61.31\) & \(27.08\) & \(88.39\) \\
        \lm{OLMo2-1124-7B}         & \(491\)   & \(-3.65\) & \(-62.77\) & \(22.41\) & \(85.18\) \\
        \lm{Gemma-3-4B-pt}         & \(534\)   & \(-6.17\) & \(-78.83\) & \(17.79\) & \(96.62\) \\
        \lm{Mistral-Nemo-2407}         & \(527\)  & \(-4.15\) & \(-71.16\) & \(17.83\) & \(88.99\) \\
        \midrule
        \multicolumn{6}{c}{\textbf{Code}} \\
        % \lm{Qwen2.5-Coder-14B}          & \(4904\)   & \(-12.53\) & \(-69.32\) \\
        \lm{Qwen3-32B}             & \( 5223\)   & \(-7.53\) & \(-87.60\) & \(5.05\) & \(88.48\) \\
        \lm{Llama-3.1-8B}  & \(5863\)   & \(-6.83\) & \(-94.03\) & \(5.20\) & \(99.23\) \\
        \lm{OLMo2-1124-7B}              & \(5863\)   & \(-6.85\) & \(-92.63\) & \(6.65\) & \(99.28\) \\
        \lm{Gemma-3-4B-pt}              & \(2325\)  & \(-2.26\) & \(-95.61\) & \(3.95\) & \(99.56\) \\
        \lm{Mistral-Nemo-2407}         & \(4729\)  & \(-6.92\) & \(-94.80\) & \(4.96\) & \(99.76\) \\
        \bottomrule
    \end{tabular}
    \label{tab:result-repeat}
\end{table*}

\subsection{Metrics}
In \autoref{subsec:prompt_construction}, we created $(\operatorname{\textcharcolor{prompt}},\operatorname{\textcharcolor{continuation}})$ pairs.
We denote $\texttokencolor{T}$ as the token that bridges the prompt-continuation boundary, with $\textcharcolor{s_1}$ being the string portion of $\texttokencolor{T}$ in the prompt and $\textcharcolor{s_2}$ being the string portion in the continuation ($\operatorname{decode}(\texttokencolor{T})=\textcharcolor{s_1} + \textcharcolor{s_2}$).
Because $\operatorname{\textcharcolor{prompt}}$ ends at a word boundary by construction, we refer to it as the \textit{word-aligned prompt}.
For comparison, we also construct a \textit{token-aligned prompt} by backing off the $\operatorname{\textcharcolor{prompt}}$ to the closest token boundary, i.e., $\operatorname{\textcharcolor{prompt}} - \textcharcolor{s_1}$ where we use $-$ to denote removal of a string suffix.

We use two metrics to evaluate a model's sensitivity to the partial token problem:
\begin{itemize}[leftmargin=12pt]
    \item \textbf{Difference in log-probability} 
    % \nascomment{consistently everywhere you say ``log probability'' but the equation below doesn't have the logs!!!} \alisa{oh crap} 
    that is assigned to the 
    % (respective) 
    correct next-token when given the word-aligned versus token-aligned prompt.
    The correct next-token when conditioning on $\operatorname{\textcharcolor{prompt}}$ is $\texttokencolor{c_1}$, i.e., the first element of $\encode(\operatorname{\textcharcolor{continuation}})=\langle \texttokencolor{c_1},...,\texttokencolor{c_n}\rangle$; when conditioning on the token-aligned prompt, the expected next-token is $\texttokencolor{T}$.
    \begin{multline*}
        \Delta \operatorname{Logprob} = \log P_M(\texttokencolor{c_1}\mid \encode(\operatorname{\textcharcolor{prompt}}))\\
        -\log P_M(\texttokencolor{T}\mid \encode(\operatorname{\textcharcolor{prompt}} - \textcharcolor{s_1}))
    \end{multline*}
    % The first term is the probability assigned to the true next-token for the realistic user prompt, whereas the second is the probability assigned to the true next-token when the prompt is backed off using knowledge of the underlying tokenization.

    \item \textbf{Difference in accuracy} of greedily decoding the correct continuation when given the word-aligned versus token-aligned prompt. %\alisa{with greedy decoding? note to self: come up with a corresponding math equation for this using argmax}
    Let $\textcharcolor{x_1}$ be the string continuation given the word-aligned prompt and $\textcharcolor{x_2}$ be that given the token-aligned prompt (in practice, we decode 3 tokens which is empirically sufficient).
    % Let \(\texttokencolor{\hat{t}_1}=\argmax_{\texttokencolor{v} \in \mathcal V} P_M(\texttokencolor{v} | \encode(\operatorname{\textcharcolor{prompt}}))\) be the most likely next-token given the word-aligned prompt and \(\texttokencolor{\hat{t}_2}=\argmax_{\texttokencolor{v} \in \mathcal V} P_M(\texttokencolor{v} | \encode(\operatorname{\textcharcolor{prompt}} - \textcharcolor{p}))\) be that given the token-aligned prompt, where \(\mathcal V\) is the vocabulary.
    Then
    \[
        \Delta \operatorname{Acc} = \frac{1}{N} \sum_{i=1}^N \mathbbm{1}[\textcharcolor{s_2}\sqsubseteq\textcharcolor{x_1}] - \frac{1}{N} \sum_{i=1}^N \mathbbm{1}[\textcharcolor{s_1} + \textcharcolor{s_2}\sqsubseteq\textcharcolor{x_2}]
    \]

    where \(N\) is the number of test instances, \(\mathbbm{1}\) is the indicator function, and \(\sqsubseteq\) denotes a string prefix.
    In other words, the generation \(\textcharcolor{x_1}\) given the word-aligned prompt should cover the remainder $\textcharcolor{s_2}$ of the truncated token \(\texttokencolor{T}\), and the generation \(\textcharcolor{x_2}\) given the token-aligned prompt should cover the entirety of \(\operatorname{decode}(\texttokencolor{T})=\textcharcolor{s_1} + \textcharcolor{s_2}\).
\end{itemize}

% \subsection{Models}

% \alisa{I'm considering not even mentioning the model names in the running text. they can read off the results table, so it feels like a waste of time.}
% We evaluate the \lm{Qwen3} and \lm{Hunyuan} pretrained series, which prioritized Chinese data in training.
% We evaluate multilingual LMs including \lm{Qwen3}, \lm{Llama-3.2}, \lm{Mistral}, and \lm{Gemma 3}. 
% We evaluate both coder models, such as Qwen3 Coder, and general-purpose models with strong coding capabilities, including Llama 3.1 and 3.2, Mistral, OLMo 2, and Gemma 3.

\subsection{Results}
% Across all evaluated models, we observe a consistent negative log-probability gap between token-aligned and work-aligned prompts, confirming that PTP problem systematically reduces both completion accuracy and prediction confidence. 

% \paragraph{Models consistently suffer from PTP.} Models consistently show low accuracy, with $\Delta$Prob being negative for all tested models, showing that PTP significantly reduces the accuracy and model confidence in predicting the true continuation of a token. 

% \paragraph{Models show low accuracy under PTP.} All models achieve surprisingly low accuracy across every domain. Even the largest models fail, such as Qwen3-32B on Chinese (7.16\% Acc.), Mistral-Nemo 12B on German(2.93\%), and Qwen3-Coder-30B on coding (1.38\%).  \hao{change to delta acc description}

\paragraph{Model performance drops substantially under PTP} 
Despite the trivial nature of the task, unexpected token boundaries compromise a serious failure mode for all models evaluated.
Across all settings, models predict the correct next-token 60\% -- 95.6\% less often and place up to four orders of magnitude less probability on the correct next-token.
Qualitatively, we observe that predicted continuations given the partial token problem often exhibit one of a few common patterns: generating an extra space (e.g., $\token{.\whitespace append}$ instead of $\token{.append}$), skipping a character (e.g.,  $\token{def \_init\_\_}$ instead of $\token{def \_\_init\_\_}$), or choosing a semantically equivalent continuation (e.g., $\token{的1/2}$ (\textit{1/2 of}) instead of $\token{的一半}$ (\textit{half of})). %\hao{color prompt vs. continuation?} \alisa{I think we should add a short parenthetical example for each case} 

\begin{figure*}[t]
    \centering
    \begin{subfigure}[t]{0.49\linewidth}
        \centering
        \includegraphics[width=\linewidth]{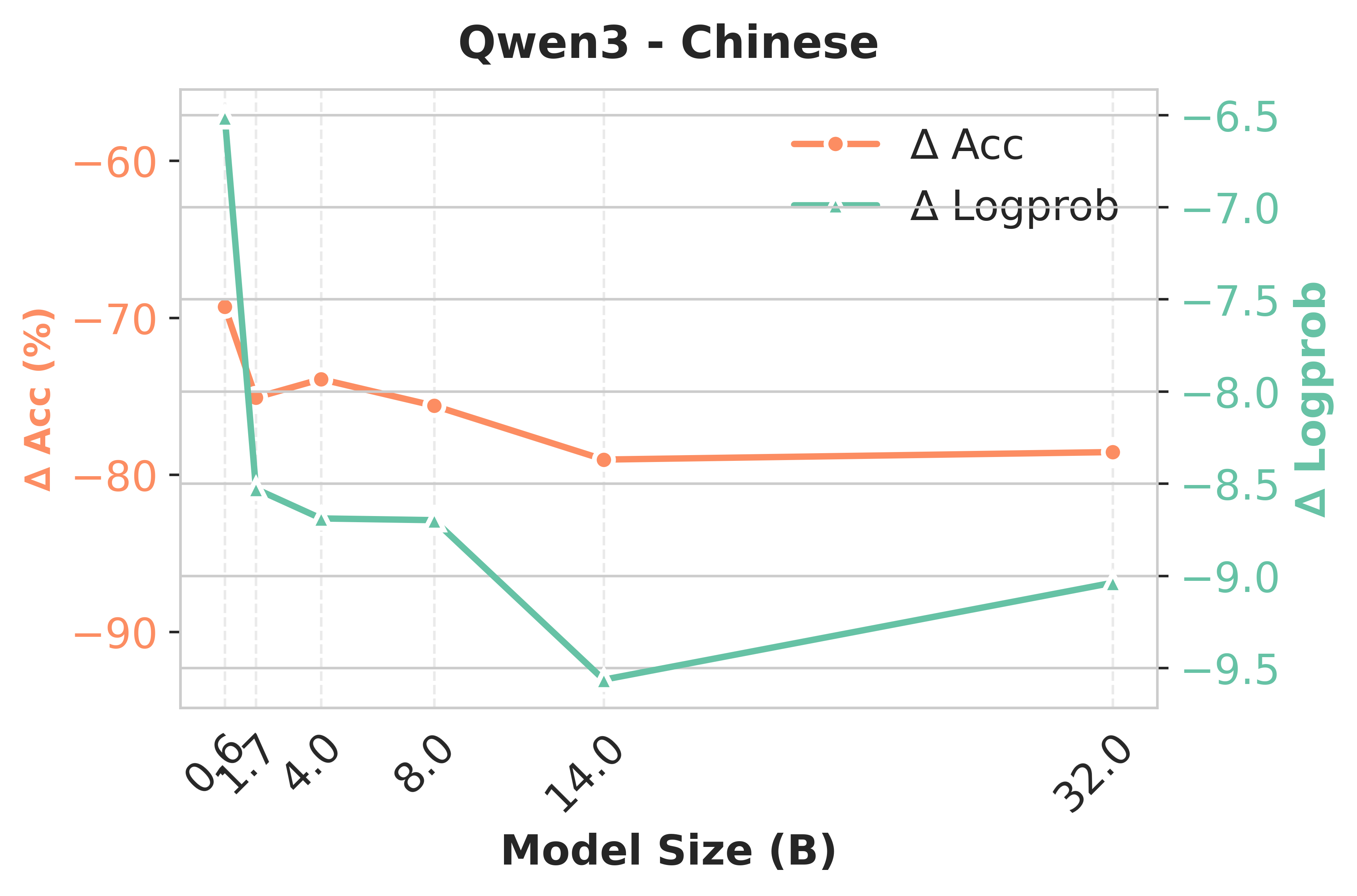}
    \end{subfigure}
    \hfill
    \begin{subfigure}[t]{0.49\linewidth}
        \centering
        \includegraphics[width=\linewidth]{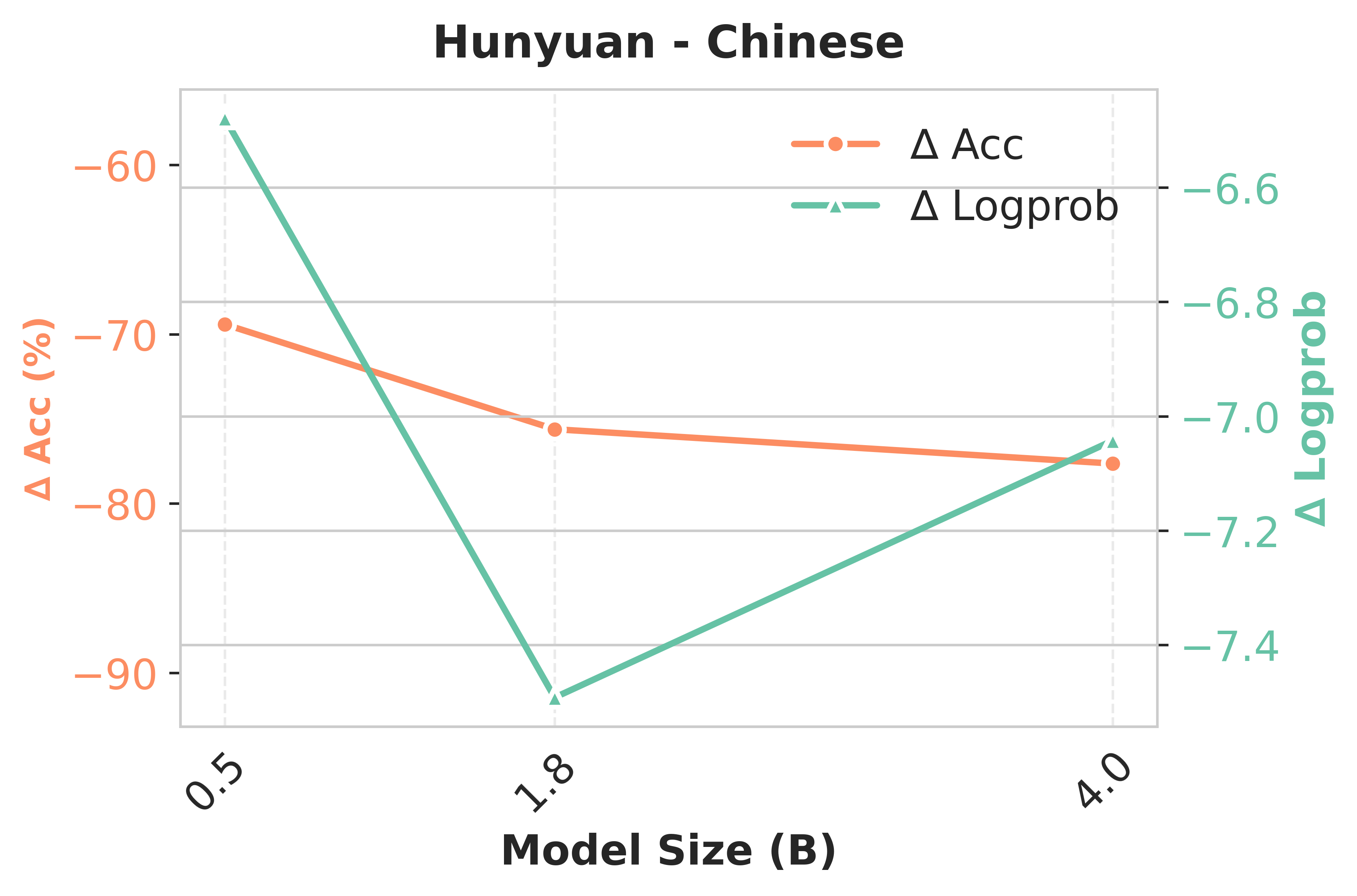}
    \end{subfigure}
    \begin{subfigure}[t]{0.49\linewidth}
        \centering
        \includegraphics[width=\linewidth]{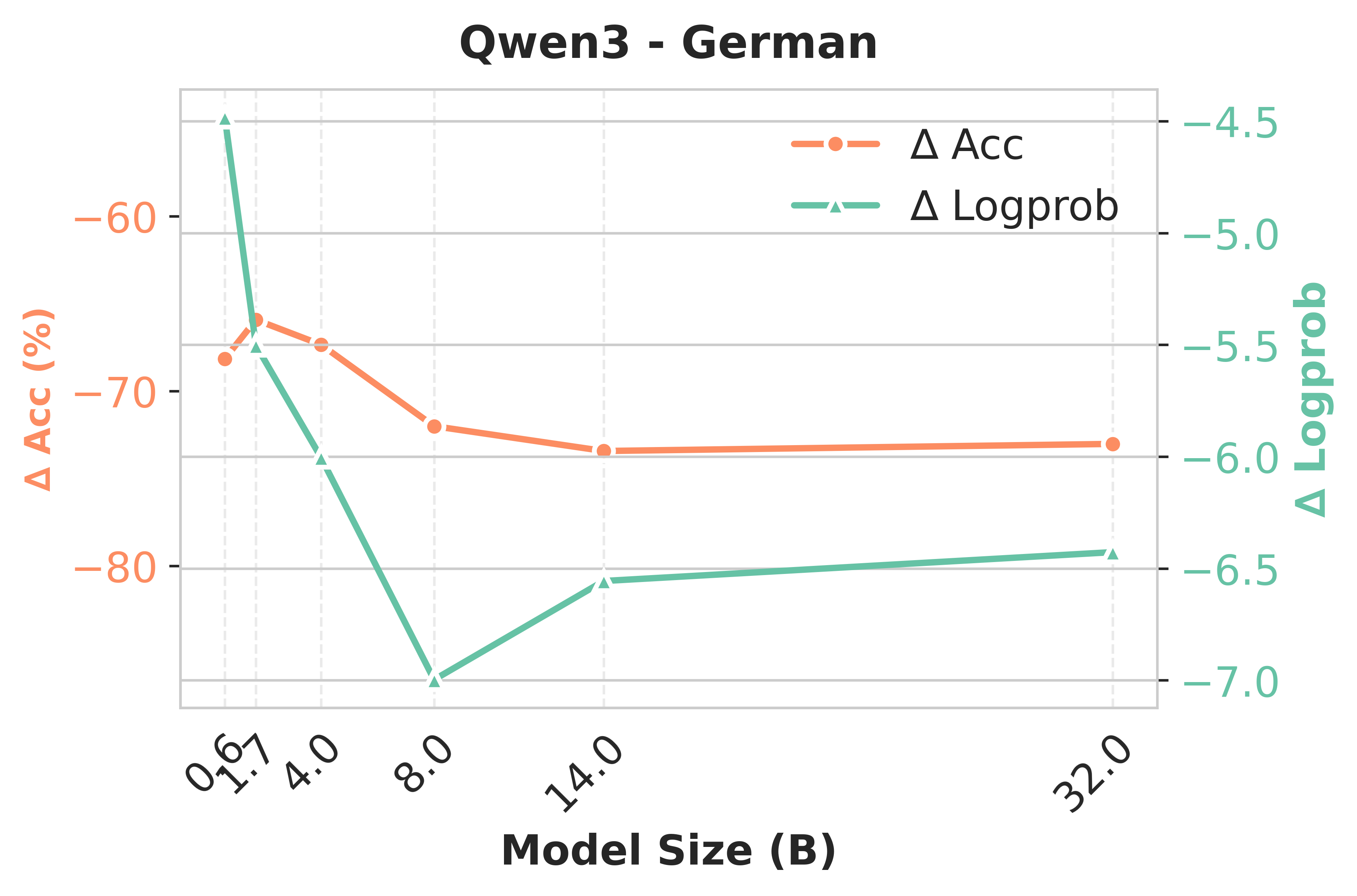}
    \end{subfigure}
    \hfill
    \begin{subfigure}[t]{0.49\linewidth}
        \centering
        \includegraphics[width=\linewidth]{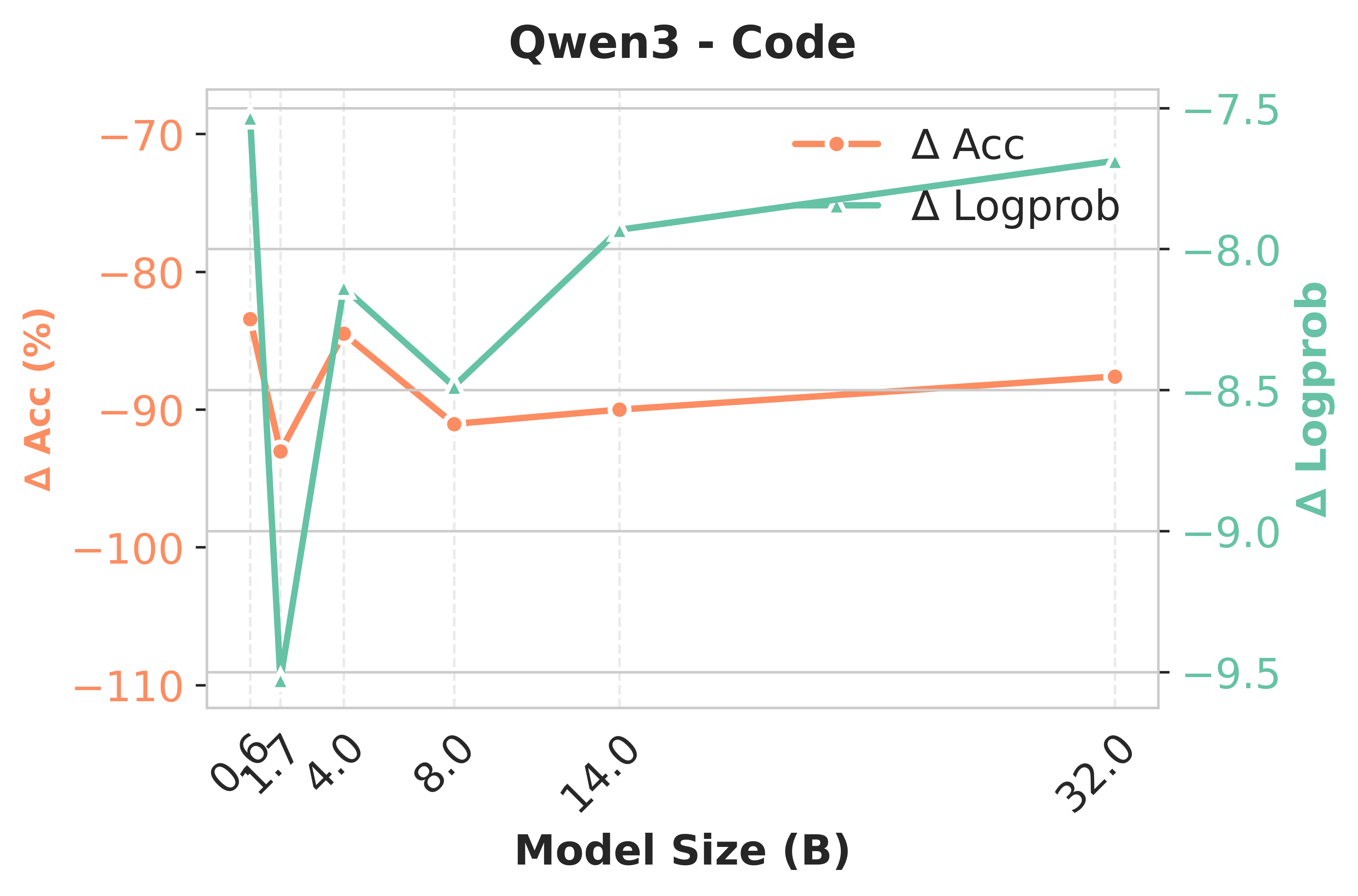}
    \end{subfigure}
    \caption{\textbf{Sensitivity to the PTP does not improve with scale across model families}. Larger models are often equally or more affected by unexpected token boundaries caused by the prompt ending.}
    % \alisa{use $\Delta$ Acc and $\Delta$ Logprob on the $y$-axes to stay consistent with the metric names introduced eralier. the bottom-right subplot uses a different color scheme for some reason :P too much spacing between the left and right column of subplots.}}
    \label{fig:scaling}
    \vspace{-1em}
\end{figure*}

\paragraph{Scaling does not mitigate the problem}
For model families where multiple model sizes are available, we plot the relationship between model size (in number of parameters) and our two metrics, \(\Delta \operatorname{Acc}\) and \(\Delta \operatorname{Logprob}\).
% \alisa{I think we should just say that it isn't improving with scale. it's not cleanly scaling inversely, either.}
Shown in \autoref{fig:scaling}, performance does not improve with scale. Larger models experience equal or greater drops in both accuracy and log probability compared to their smaller counterparts.
% there is an inverse scaling trend \citep{mckenzie-etal-2023-inverse} along both metrics for all model families: the performance degradation caused by the partial token problem is larger in larger models.

We hypothesize that robustness to partial tokens does not improve with scale because larger models are, in fact, better fit to the tokenizer that was used in training.
Given the context $\langle\token{\whitespace process}, \token{in}\rangle$, it is more confident that the next token is \textit{not} \token{g} because the sequence $\langle\token{\whitespace process}, \token{in}, \token{g}\rangle$ was never seen in training (it cannot be produced by the tokenizer; the token \token{\whitespace processing} would have observed instead).
This indicates that scaling cannot be relied on to resolve this issue, and in the next section, we instead explore the effectiveness of inference-time solutions.

% When models fail to generate correctly in \textit{word-aligned} case, we qualitatively observe the model outputs 

% We qualitatively observe that the LM often "gets around" the partial token problem by generating an extra space, skipping a character, or choosing a semantically equivalent alternative next-character.

\section{Exploring inference-time solutions}\label{subsec:bytesampler}

Are inference-time approaches effective for overcoming the PTP?
We consider a heuristic approach, token healing \citep{ribeiro2023guidance}, and an exact approach, \texttt{ByteSampler} \citep{hayase-etal-2025-sampling}.
While token healing backs off the prompt by exactly one token, in general it is not known in advance how far to back off. \texttt{ByteSampler} therefore constructs a \textit{tree} of all (valid) token sequences that cover the prefix.
Paths in this tree (from root to leaf) correspond to token sequences consistent with the prompt, and are not constrained to having a token boundary at the end of the prompt.
After this, we can extend the token sequence by sampling normally, since the sequence ends on a token selected by the model.

\subsection{Setup}
% In \autoref{sec:experiments}, we saw that the \textit{word-aligned} prompt (which ends with a complete word but partial token) consistently produces worse performance than the \textit{token-aligned} counterpart (which backs off to the nearest token boundary).
% Here, we test whether infe with the word-aligned prompt will recover the desired performance.
Like in \autoref{sec:experiments}, we consider the prediction accurate if the decoded byte string covers the remainder $\textcharcolor{s_2}$ of the truncated token $\texttokencolor{T}$.
We omit the probability-based metric as calculating it is more complex under the two methods.
Note that applying token healing to the word-aligned prompt differs from the token-aligned prompt (which is also backed off) in two ways: (1) they do not necessarily correspond to the same prompt ($\textcharcolor{s_1}$ is not necessarily one token), and (2) token healing additionally constrains the generated continuation to match the backed-off text.

\subsection{Results}

Results for \lm{Qwen3-32B} are shown in \autoref{tab:bytesampler}; please see results for additional models in \autoref{app:add-inference-result}.
While the effectiveness of token healing is mixed, we find that invoking \texttt{ByteSampler} on the word-aligned prompt, strikingly, achieves 100.00\% accuracy across all three domains.
The stronger performance over the token-aligned setting makes sense, as the word-aligned prompt provides strictly stronger conditioning (it is longer by at least one additional word) while asking for the completion of the same string.
\texttt{ByteSampler} also uses requires low overhead, requiring on average 0.12 -- 1.17 additional forward passes compared to regular sampling.

\begin{table*}[h]
    \centering
    \caption{\textbf{Token healing and \texttt{ByteSampler} both mitigate the partial token problem at inference-time.}
    Token healing shows mixed effectiveness, while \texttt{ByteSampler} overcomes the partial token problem completely.
    \textbf{Acc (\%)} is exact-match continuation accuracy; \textbf{Overhead} is the average number of additional forward passes over normal sampling. Results are for \lm{Qwen3-32B}.}
    \begin{tabular}{lrrr}
        \toprule
        \textbf{Setting} & \textbf{Acc} (\%) & $\Delta$ & \textbf{Overhead} \\
        \midrule
        \multicolumn{4}{c}{\textbf{Chinese}} \\
        Token-aligned                & 93.92 & 0.00   & 0 \\
        Word-aligned                 & 15.36 & -78.56 & 0 \\
        \quad + Token healing        & 99.08 & +5.16  & 0 \\
        \quad + \texttt{ByteSampler} & \textbf{100.00} & +6.08  & 0.65 \\
        \midrule
        \multicolumn{4}{c}{\textbf{German}} \\
        Token-aligned                & 89.45 & 0.00   & 0 \\
        Word-aligned                 & 16.43 & -73.02 & 0 \\
        \quad + Token healing        & 83.16 & -6.29  & 0 \\
        \quad + \texttt{ByteSampler} & \textbf{100.00} & +10.55 & 1.17 \\
        \midrule
        \multicolumn{4}{c}{\textbf{Code}} \\
        Token-aligned                & 88.48 & 0.00   & 0 \\
        Word-aligned                 & 5.05  & -83.43 & 0 \\
        \quad + Token healing        & 94.09 & +5.61  & 0 \\
        \quad + \texttt{ByteSampler} & \textbf{100.00} & +11.52 & 1.12 \\
        \bottomrule
    \end{tabular}
    \label{tab:bytesampler}
    \vspace{-1em}
\end{table*}

\section{Related work and discussion}

% \paragraph{Tokenization in Language Models.}  
% Tokenization is a core preprocessing step in LM, enabling the mapping of raw text into understandable units. Popular methods include Byte-Pair Encoding (BPE) \citep{sennrich2016neuralmachinetranslationrare}, WordPiece \citep{wu2016googlesneuralmachinetranslation}, and Unigram \citep{kudo2018subwordregularizationimprovingneural} balance vocabulary efficiency and open-vocabulary handling. While these techniques are well-optimized for English and morphologically simple languages, their limitations become more apparent in multilingual settings, especially in logographic or compounding languages.

\paragraph{Partial token problem}
While the PTP is well-known in the literature, its impact is generally not felt in English left-to-right generation, nor in chat applications where special tokens separate the prompt and response.
Nonetheless, we believe that a better understanding of and robust solution to the PTP will enable more future possibilities in tokenization.
For instance, recent ``superword'' tokenization methods \citep{liu-etal-2025-superbpe, schmidt-etal-2025-boundless}, which improve compression by including multi-word tokens, may encounter the PTP even in English use.

Methods proposed to address the PTP so far (discussed in \autoref{sec:background}) all operate at sampling-time, without requiring changes to model training.
However, another possible approach is using \textit{stochastic tokenization} in training \citep{provilkov-etal-2020-bpe, sims-etal-2025-stochastok, cognetta-etal-2024-distributional} to expose the model to multiple possible segmentations of the same text.
These methods are mainly motivated by improving models' subword understanding, however, and its utility for addressing the PTP remains unknown.
Nonetheless, \lm{Deepseek v3} \citep{deepseek-2025-deepseekv3} hints at this possibility: to address the PTP, it randomly splits some proportion of multi-punctuation tokens into smaller tokens in training, though experiments are not presented with this ablation.
We note that stochastic tokenization more broadly would reduce the tokenizer's encoding efficiency and may have other side effects, such as diluting the probability of a string over many different tokenizations \citep{song-etal-2024-submerge}.
We thus recommend pairing inference-time fixes with the current standard of deterministic tokenization.

\paragraph{Token-word misalignment}  
It is well-known that tokenizers, especially when learned with the BPE algorithm \citep{sennrich-etal-2016-neural}, generally do not respect derivational, compound, or morphological boundaries within words that are apparent to humans \citep{chai-etal-2024-tokenization, minixhofer-etal-2023-compound}.
This has motivated many efforts towards linguistically-informed tokenization \citep{klein-tsarfaty-2020-getting, hofmann-etal-2021-superbizarre, hofmann-etal-2022-embarrassingly, yehezkel-pinter-2023-incorporating, bauwens-delobelle-2024-bpe, si-etal-2023-sub}.
% \alisa{weird transition here}
% In contrast, we focus on how this leads to the PTP in natural use cases.
In this work, we highlight a practical consequence of this misalignment: realistic user prompts can yield distorted next-token distributions and unexpected continuations.

\paragraph{Robustness to tokenization}
Our work also adds to discussion on the limitations of using tokenization.
It is commonly argued that tokenization obscures orthographic information of tokens \citep{edman-etal-2024-cute, chai-etal-2024-tokenization, wang-etal-2024-tokenization}, though other evidence suggests that LMs naturally learn the characters that make up tokens, especially at scale \citep{kaushal-mahowald-2022-tokens, itzhak-levy-2022-models, feucht-etal-2024-token, kaplan-etal-2025-tokens}.
Recent related work also shows that models retain semantic understanding of non-canonical tokenizations of the context \citep{zheng-etal-2025-broken, geh-etal-2025-adversarial, kaplan-etal-2025-tokens}, e.g., recognizing that the character-level tokenization $\langle\token{\whitespace}, \token{c}, \token{a}, \token{t}\rangle$ corresponds to \chars{\whitespace cat}.
Note that this finding is complementary with ours: one is about understanding non-canonical tokenizations in the context history, whereas our work is about \textit{generating} non-canonical tokenizations.

\section{Conclusion}

Language models are designed to provide probability distributions over strings, but are engineered to learn distributions over sequences of tokens.
As a result, they model not only language but also the tokenizer that they are trained with.
In this work, we revisit a fundamental problem that ensues: the ending of a user-provided prompt forces a token boundary that biases the prediction of the model.
We demonstrate that this problem appears commonly in natural use cases in languages where whitespace does not reliably delimit semantic units, and that this significantly degrades generated continuations.
We then provide recommendations for inference-time solutions by evaluating them over our test set.
We hope that our work deepens understanding of the interaction between tokenization and language modeling.


\clearpage
\appendix

\section*{Impact Statement}
This work studies a failure mode in language model inference caused by mismatches between tokenization and natural language. By characterizing the prevalence and severity of the partial token problem in realistic settings and evaluating inference-time mitigations, our results aim to improve the reliability and correctness of deployed language models. We do not anticipate negative societal impacts beyond those already associated with language model deployment. 


\bibliography{paper}
\bibliographystyle{icml2026}

%%%%%%%%%%%%%%%%%%%%%%%%%%%%%%%%%%%%%%%%%%%%%%%%%%%%%%%%%%%%%%%%%%%%%%%%%%%%%%%
%%%%%%%%%%%%%%%%%%%%%%%%%%%%%%%%%%%%%%%%%%%%%%%%%%%%%%%%%%%%%%%%%%%%%%%%%%%%%%%
% APPENDIX
%%%%%%%%%%%%%%%%%%%%%%%%%%%%%%%%%%%%%%%%%%%%%%%%%%%%%%%%%%%%%%%%%%%%%%%%%%%%%%%
%%%%%%%%%%%%%%%%%%%%%%%%%%%%%%%%%%%%%%%%%%%%%%%%%%%%%%%%%%%%%%%%%%%%%%%%%%%%%%%
\newpage
\appendix
\onecolumn
\section{Alternative Experimental Setup}
\label{app:translation}

Following the same method of prompt construction as in \autoref{subsec:prompt_construction}, we create another set of tasks with details described below.

\paragraph{Chinese}
We frame the task as translation from English to Chinese using parallel corpus FLORES.
Chinese sentences are segmented with \texttt{Jieba} for words and the model's tokenizer for tokens. 
Prompts present the English source and request a Chinese translation, with the target translation partially provided. We use the prompt
``\textit{Translate the following sentence from English to Chinese: \newline English: \{English sentence\} \newline Chinese: \{prefix\} }''.

\paragraph{German}
We follow the same approach as for Chinese: we frame the task as 
translation from English to German, 
use the parallel corpus FLORES, and use the segmentor \texttt{CharSplit}.

\paragraph{Code}
The prompt consists of the docstring, which specifies the desired functionality, followed by a prefix from the solution code that ends within multiple-character punctuation tokens. We use the prompt
``\textit{Complete the code based on the instruction: \newline Instruction: \{docstring\} \newline Code: \{prefix\}}''.

Results are shown in \autoref{tab:translate-results} and support the same findings as our main experiments.
We see that when facing the PTP, models place at least one order of magnitude less probability on the true continuation and accuracy of model prediction drops by at least 30\%.

\begin{table}[h]
\centering\small
\setlength{\tabcolsep}{6pt}
\renewcommand{\arraystretch}{1.2}
\begin{tabular}{lrrrrr}
    \toprule
    \textbf{Model} & $N$ & \textbf{\(\Delta\) Logprob} & \textbf{\(\Delta\) Acc (\%)} & \textbf{Acc$_{\text{word}}$ (\%)} & \textbf{Acc$_{\text{token}}$ (\%)} \\
    \midrule
    \multicolumn{6}{c}{\textbf{Chinese}} \\
    \lm{Qwen3-32B} & \(2304\) & \(-9.59\) & \(-34.98\) & \(15.97\) & \(50.95\) \\
    \lm{Hunyuan-4B-Pretrain}   & \(3551\) & \(-7.92\) & \(-30.75\) & \(20.58\) & \(51.33\) \\
    \midrule
    \multicolumn{6}{c}{\textbf{German}} \\
    \lm{Qwen3-32B}             & \(493\)   & \(-7.16\) & \(-40.16\) & \(15.41\) & \(55.57\) \\
    \lm{Llama-3.1-8B}          & \(513\)   & \(-4.75\) & \(-33.92\) & \(18.12\) & \(52.04\) \\
    \lm{Mistral-Nemo-2407}     & \(527\)   & \(-4.01\) & \(-48.58\) & \(17.31\) & \(49.69\) \\
    \lm{OLMo2-1124-7B}         & \(491\)   & \(-3.44\) & \(-32.38\) & \(18.21\) & \(66.79\) \\
    \lm{Gemma-3-4B-pt}         & \(534\)   & \(-4.75\) & \(-45.13\) & \(15.91\) & \(61.04\) \\
    \midrule
    \multicolumn{6}{c}{\textbf{Code}} \\
    \lm{Qwen3-32B}             & \( 5223\)   & \(-8.64\) & \(-65.82\) & \(5.20\) & \(71.02\) \\
    \lm{Llama-3.1-8B}  & \(5863\)   & \(-6.66\) & \(-65.76\) & \(5.26\) & \(71.02\) \\
    \lm{Mistral-Nemo-2407}         & \(4729\)  & \(-6.81\) & \(-71.05\) & \(5.13\) & \(76.18\) \\
    \lm{OLMo2-1124-7B}              & \(5863\)   & \(-5.88\) & \(-58.62\) & \(6.24\) & \(64.86\) \\
    \lm{Gemma-3-4B-pt}              & \(2325\)  & \(-0.77\) & \(-33.94\) & \(1.67\) & \(35.61\) \\
    \bottomrule
\end{tabular}
\caption{All models suffer a significant drop in prediction accuracy and probability on the correct continuation although the continuation is less unambiguous.}
\label{tab:translate-results}
\vspace{-2em}
\end{table}

\clearpage

\section{Additional Inference-Time Solution Results}
\label{app:add-inference-result}
In this section, we present inference-time results for PTP on additional models, complementing the results in Table \ref{tab:result-repeat}.

\begin{table}[h]
\centering\small
\caption{Inference-time mitigations for the partial token problem on \textbf{Chinese}.}
\begin{tabular}{lrrr}
\toprule
\textbf{Setting} & \textbf{Acc} (\%) & $\Delta$ & \textbf{Overhead} \\
\midrule
\multicolumn{4}{c}{\lm{Hunyuan-4B-Pretrain}} \\
Token-aligned                & 94.00 & 0.00   & 0 \\
Word-aligned                 & 16.36 & -77.64 & 0 \\
\quad + Token healing         & 97.90 & +3.90  & 0 \\
\quad + \texttt{ByteSampler} & \textbf{100.00} & +6.00  & 1.38 \\
\bottomrule
\end{tabular}
\label{tab:add-zh}
\vspace{-1em}
\end{table}

\begin{table}[h]
\centering\small
\caption{Inference-time mitigations for the partial token problem on \textbf{German}.}
\begin{tabular}{lrrr}
\toprule
\textbf{Setting} & \textbf{Acc} (\%) & $\Delta$ & \textbf{Overhead} \\
\midrule
\multicolumn{4}{c}{\lm{Llama-3.1-8B}} \\
Token-aligned                & 88.39 & 0.00   & 0 \\
Word-aligned                 & 27.08 & -61.31 & 0 \\
\quad + Token healing        & 84.99 & -3.40  & 0 \\
\quad + \texttt{ByteSampler} & \textbf{100.0}0 & +11.61 & 0.17 \\
\midrule
\multicolumn{4}{c}{\lm{OLMo2-1124-7B}} \\
Token-aligned                & 85.18 & 0.00   & 0 \\
Word-aligned                 & 22.41 & -62.77 & 0 \\
\quad + Token healing        & 83.70 & -1.48  & 0 \\
\quad + \texttt{ByteSampler} & \textbf{100.00} & +14.82 & 1.17 \\
\midrule
\multicolumn{4}{c}{\lm{Mistral-Nemo-2407}} \\
Token-aligned                & 88.99 & 0.00   & 0 \\
Word-aligned                 & 17.83 & -71.16 & 0 \\
\quad + Token healing        & 90.70 & +1.71  & 0 \\
\quad + \texttt{ByteSampler} & \textbf{100.00} & +11.01 & 0.41 \\
\bottomrule
\end{tabular}
\label{tab:add-de}
\vspace{-1em}
\end{table}

\begin{table}[h]
\centering\small
\caption{Inference-time mitigations for the partial token problem on \textbf{Code}.}
\begin{tabular}{lrrr}
\toprule
\textbf{Setting} & \textbf{Acc} (\%) & $\Delta$ & \textbf{Overhead} \\
\midrule
\multicolumn{4}{c}{\lm{Llama-3.1-8B}} \\
Token-aligned                & 99.23 & 0.00   & 0 \\
Word-aligned                 & 5.20  & -94.03 & 0 \\
\quad + Token healing        & 99.65 & +0.42  & 0 \\
\quad + \texttt{ByteSampler} & \textbf{100.00} & +0.77  & 0 \\
\midrule
\multicolumn{4}{c}{\lm{OLMo2-1124-7B}} \\
Token-aligned                & 85.18 & 0.00   & 0 \\
Word-aligned                 & 22.41 & -62.77 & 0 \\
\quad + Token healing        & 99.13 & +13.95 & 0 \\
\quad + \texttt{ByteSampler} & \textbf{100.00} & +14.82 & 1.12 \\
\midrule
\multicolumn{4}{c}{\lm{Mistral-Nemo-2407}} \\
Token-aligned                & 99.76 & 0.00   & 0 \\
Word-aligned                 & 4.96  & -94.80 & 0 \\
\quad + Token healing        & 99.57 & -0.19  & 0 \\
\quad + \texttt{ByteSampler} & \textbf{100.00} & +0.24  & -0.93 \\
\bottomrule
\end{tabular}
\label{tab:add-code}
\vspace{-1em}
\end{table}

%%%%%%%%%%%%%%%%%%%%%%%%%%%%%%%%%%%%%%%%%%%%%%%%%%%%%%%%%%%%%%%%%%%%%%%%%%%%%%%
%%%%%%%%%%%%%%%%%%%%%%%%%%%%%%%%%%%%%%%%%%%%%%%%%%%%%%%%%%%%%%%%%%%%%%%%%%%%%%%
\end{CJK*}
\end{document}